\documentclass[11pt,a4paper]{article}
\usepackage[hyperref]{emnlp-ijcnlp-2019}
\usepackage{times}
\usepackage{amsmath}
\usepackage{graphicx}
\usepackage{caption}
\usepackage{subcaption}
\usepackage{float}
\usepackage{booktabs}
\usepackage{natbib}
\usepackage{latexsym}
\usepackage{todonotes}
\usepackage{algorithm}
\usepackage{algorithmicx}
\usepackage[noend]{algpseudocode}

\usepackage{amsmath,amsfonts,bm}

\def\eqref#1{equation~\ref{#1}}

\def\1{\bm{1}}

\DeclareMathAlphabet{\mathsfit}{\encodingdefault}{\sfdefault}{m}{sl}
\SetMathAlphabet{\mathsfit}{bold}{\encodingdefault}{\sfdefault}{bx}{n}

\usepackage{url}

\aclfinalcopy 

\title{Multi-hop Question Answering via \\  Reasoning Chains}

\author{Jifan Chen, Shih-ting Lin \and Greg Durrett \\
  The University of Texas at Austin \\
  {\tt \{jfchen,j0717lin,gdurrett\}@cs.utexas.edu}}

\date{}

\begin{document}
\maketitle
\begin{abstract}
Multi-hop question answering requires models to gather information from different parts of a text to answer a question. Most current approaches learn to address this task in an end-to-end way with neural networks, without maintaining an explicit representation of the reasoning process. We propose a method to extract a discrete reasoning chain over the text, which consists of a series of sentences leading to the answer. We then feed the extracted chains to a BERT-based QA model ~\citep{devlin2018bert} to do final answer prediction. Critically, we do not rely on gold annotated chains or ``supporting facts'': at training time, we derive pseudogold reasoning chains using heuristics based on named entity recognition and coreference resolution. Nor do we rely on these annotations at test time, as our model learns to extract chains from raw text alone.  We test our approach on two recently proposed large multi-hop question answering datasets: WikiHop~\citep{welbl2018constructing} and HotpotQA~\citep{yang2018hotpotqa}, and achieve state-of-art performance on WikiHop and strong performance on HotpotQA. Our analysis shows properties of chains that are crucial for high performance: in particular, modeling extraction sequentially is important, as is dealing with each candidate sentence in a context-aware way. Furthermore, human evaluation shows that our extracted chains allow humans to give answers with high confidence, indicating that these are a strong intermediate abstraction for this task.
\end{abstract}

\section{Introduction}

As high performance has been achieved in simple question answering settings \citep{rajpurkar2016squad}, work on question answering has increasingly gravitated towards questions that require more complex reasoning to solve. Multi-hop question answering datasets explicitly require aggregating clues from different parts of some given documents \citep{dua2019drop, welbl2018constructing, yang2018hotpotqa, jansen2018worldtree, khashabi2018looking}. Earlier question answering datasets contain some questions of this form \citep{richardson2013mctest,lai2017race}, but typically exhibit a limited range of multi-hop phenomena. Designers of multi-hop datasets aim to test a range of reasoning types \citep{yang2018hotpotqa} and, ideally, systems should have to behave in a very specific way in order to do well. However, \newcite{chen2019understanding} and \newcite{min2019compositional} show that models achieving high performance may not actually be performing the expected kinds of reasoning. Partially this is due to the difficulty of evaluating intermediate model components such as attention~\citep{jain2019attention}, but it also suggests that models may need inductive bias if they are to solve this problem ``correctly.''

\begin{figure*}[t]
\centering
\includegraphics[width=150mm,trim=7mm 0 0 0]{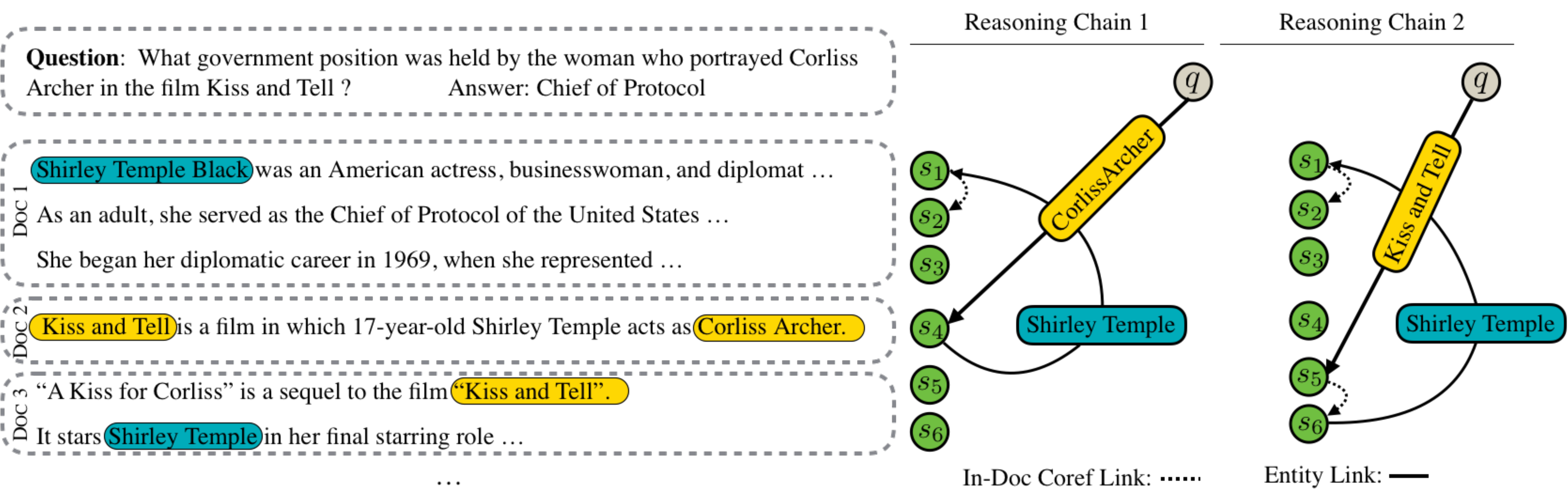}
\caption{A multi-hop example chosen from the HotpotQA development set. Several documents are given as context to answer a question. We show two possible ``reasoning chains'' that leverage connections (shared entities or coreference relations) between sentences to arrive at the answer. The first chain is most appropriate, while the second requires a less well-supported inferential leap.}
\vspace{-0.5cm}
    \label{fig:overview}
\end{figure*}

In this work, we propose a step in this direction, with a two-stage model that identifies intermediate \emph{reasoning chains} and then separately determines the answer. A reasoning chain is a sequence of sentences that logically connect the question to a fact relevant (or partially relevant) to giving a reasonably supported answer. Figure~\ref{fig:overview} shows an example of what such chains look like. Extracting chains gives us a discrete intermediate output of the reasoning process, which can help us gauge our model's behavior beyond just final task accuracy. Formally, our extractor model scores sequences of sentences and produces an $n$-best list of chains via beam search.

To find the right answer, we need to maintain uncertainty over this chain set, since the correct one may not immediately be evident, and for certain types of questions, information across multiple chains may even be relevant. Sifting through the retrieved information to actually identify the answer requires deeper, more expensive computation. We employ a second-stage answer module, a BERT-based QA system~\citep{devlin2018bert}, which can be run cheaply given the pruned context. Our approach resembles past models for coarse-to-fine question answering~\citep{choi2017coarse, min2018efficient, wang2019evidence}, but explores the context in a sequential fashion and is trained to produce principled chains.

To train our model, we heuristically label examples with reasoning chains. We use a search procedure leveraging coreference and named entity recognition (NER) to find a path from the start sentence to an end sentence through a graph of related sentences. Constructing this graph requires running an NER system at train time, but does not rely on the answer or answer candidates \citep{kundu2018exploiting}. Our system also does not require these annotations at test time, operating instead from raw text.

Our chain identification is effective and flexible: we can use it to derive supervision on two existing datasets. On HotpotQA \citep{yang2018hotpotqa}, we found that these derived chains are essentially as effective as the ground-truth supporting fact provided by the dataset. In terms of final question answering accuracy, on the WikiHop dataset \citep{welbl2018constructing}, our approach achieves state-of-the-art performance by a substantial margin, and on HotpotQA, we achieve strong results and outperform several recent published systems.

Our contributions are as follows: (1) We present a method for extracting oracle reasoning chains for multi-hop reasoning tasks. These chains generalize across multiple datasets and are comparable to human-annotated chains. (2) We present a model that learns from these chains at train time and at test time can produce a list of chains. Those chains could be used to gauge the behaviors of our model. (3) Results on two large datasets show strong performance of our chain extraction and show that the extracted chains are intrinsically a good representation of evidence for question answering.

\section{Question Answering via Chain Extraction}

We describe our notion of chain extraction in more detail. A reasoning chain is a sequence of sentences that logically connect the question to a fact relevant to determining the answer. Two adjacent sentences in a reasoning chain should be intuitively related: they should exhibit a shared entity or event, temporal structure, or some other kind of textual relation that would allow a human reader to connect the information they contain.

Figure~\ref{fig:overview} shows an example of possible reasoning chains of an real example. In this case, we need to find information about the actor who played \emph{Corliss Archer} in \emph{Kiss and Tell}. These question entities may appear in multiple places in the text, and it is generally difficult to know which entity mentions might eventually lead to text containing the answer. If we arrive at $s_4$ and find the new entity \emph{Shirley Temple}, we then need to determine what government position she held, which in this case can be found by two additional steps. Other reasoning chains could theoretically lead to this answer, such as the second chain: Shirley Temple starred in the sequel to \emph{Kiss and Tell}, which might lead us to infer that Shirley Temple also plays Corliss Archer in Kiss and Tell. Although less justified, we also view this as a valid reasoning chain. However, in general, there are also ``connected'' sequences of sentences that don't imply the answer; for example, they are connected by an entity which is not related to the question.

In determining this chain, we largely used information about entity coreference to connect the relevant pieces: either cross-document coreference about \emph{Shirley Temple} or resolution of various pronouns. Another relevant cue is that subsequent information about \emph{Shirley Temple} in Document 1 occurs later in the discourse, which in this case reflects temporal structure. However, solving coreference or temporal relation extraction in general is neither necessary nor sufficient to do chain extraction. Therefore, we design our system so that it does not rely on coreference at test time, but can instead directly extract reasoning chains based on what it has learned at training time.

Having established this notion of a reasoning chain, we have three questions to answer. First, how can we automatically select pseudo-ground-truth reasoning chains? Second, how do we model the chain extraction process? Third, how do we take one or more extracted chains and turn them into a final answer? We answer these three questions in the next section.

\section{Learning to Extract Chains}

\subsection{Heuristic oracle chain construction}

Following the intuition in Figure~\ref{fig:overview}, we assume that there are two relevant connections between sentences that can form reasoning chains. First, the presence of a shared entity often implies some kind of connection. This is not always a sufficient clue, since common entities like \emph{United States} may occur in otherwise unrelated sentences; however, because this oracle is only used at train time, it does not need to be 100\% reliable for the model to learn a chain extraction procedure. Second, we assume that any two sentences in the same paragraph are connected; this is often true on the basis of coreference or other kinds of bridging anaphora.

We derive heuristic reasoning chains by searching over a graph which is constructed based on these factors. Each sentence $s_i$ is represented as a node $i$ in the graph. We run an off-the-shelf named entity recognition system to extract all entities for each sentence. If sentence $i$ and sentence $j$ contain a shared entity, we add an edge between node $i$ and $j$. We then also add an edge between all pairs of sentence within the same paragraph.\footnote{We do not explicitly run a coreference system here since current coreference systems often introduce spurious arcs. Moreover, cross-document links can nearly always be found by exact string match, and since we add all within-paragraph links, exactly determining the coreference status of every mention is not needed.}

Starting from the question node, we do an exhaustive search to find all possible chains that could lead to the answer. This process yields a set of possible chains with different lengths; two examples are shown in Figure~\ref{fig:overview}. We use two different criteria to select heuristic oracles:
\begin{itemize}
    \item \textbf{Shortest Path}: We simply take the shortest chain from the chain set as our oracle.
    \item \textbf{Question Overlap}: We compute the \texttt{Rouge-F1} score for each chain's sentences with respect to the question and take the chain with the highest score. This encourages selection of more complete answer chains which address all of the question's parts without finding shortcuts. 
\end{itemize}



\subsection{Chain extraction model}

\begin{figure*}[t]
\centering
\includegraphics[width=150mm,trim=7mm 0 0 0]{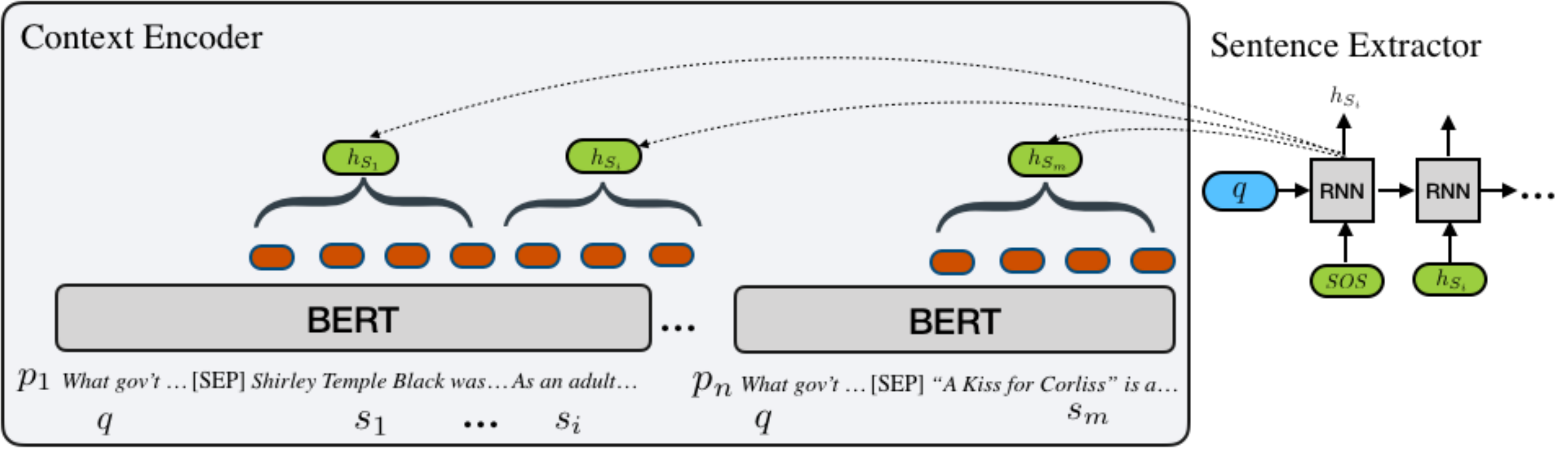}
\caption{The BERT-Para variant of our proposed chain extractor. Left side: we encode each document paragraph jointly with the question and use pooling to form sentence representations. Right side: we use a pointer network extracts a sequence of sentences.}
\vspace{-0.5cm}
    \label{fig:model_architecture}
\end{figure*}

Our chain extractor takes the input documents and questions as input and returns a variable-length sequence of sentence pointers as output. The processing flow of our chain extractor can be divided into two main parts: sentence encoding and chain prediction as shown in Figure~\ref{fig:model_architecture}.

\paragraph{Sentence Encoding}
Given a document containing $n$ paragraphs and a question, we first concatenate the question with each paragraph and then encode them using the pre-trained BERT encoder~\citep{devlin2018bert}. We denote the encoded $i$th paragraph as $p_i$. We also encode the question by itself with BERT, denoting as $q$. To compute the representation of a sentence, we extract it from the encoded paragraph. Suppose sentence $j$ in the document is the $j$th sentence of paragraph $i$. Then $\mathbf{s}_j = \text{Span\_Extractor}(p_i, s^{\text{START}}_j, s^{\text{END}}_j )$. For simplicity, we choose max-pooling as our span extractor, though other choices are possible. We name this scheme of sentence representation as \texttt{BERT-Para}. This paragraph-factored model is much more efficient and scalable than attempting to run BERT on the full context, as full contexts can be thousands of words long. We also explore an even more factored version where each sentence is concatenated with the question and encoded independently, which we denote as \texttt{BERT-Sent}. Finally, instead of using BERT as the sentence encoder, we use a bidirectional attention layer between the passage and question~\citep{seo2016bidirectional} as a baseline; we call this model \texttt{BiDAF-Para}.

\paragraph{Chain Prediction}
We treat all the encoded sentence representations as a bag of sentences and adopt an LSTM-based pointer network~\citep{vinyals2015pointer} to extract the reasoning chain, shown on the right side of Figure~\ref{fig:model_architecture}. At the first time step, we initialize the hidden state $\mathbf{h}_0$ in the pointer network using the max-pooled representation of the question $q$, and feed a special token \texttt{SOS} as the first input.

Let $c_1,\ldots,c_l$ denote the indices of sentences to include in the reasoning chain. At time step $t$, we compute the probability of sentence $i$ being chosen as $P(c_t=i|c_1,\ldots,c_{t-1},\mathbf{s}) = \textrm{softmax}(\mathbf{\alpha})[i]$, where $\alpha_i = \mathbf{W} [\mathbf{h}_{t-1}; \mathbf{s}_{c_{t-1}}; \mathbf{h}_{t-1}\odot \mathbf{s}_{c_{t-1}}]$, and $\mathbf{W}$ is a weight matrix to be learned.


\paragraph{Training the Chain Extractor} During training, the loss for time step $t$ is the negative log likelihood of the target sentence $c^*_t$ for that time step: $\textrm{loss}_t = -\log(P(c_t^*)|c_1^*,\ldots,c_{t-1}^*\mathbf{s})$. We also explored training with reinforcement learning. For the two datasets we considered, pre-training with our oracle and fine-tuning with policy gradient this did not lead to an improvement. Pure oracle chain extraction appears strong enough for the model to learn the needed associations across chain timesteps, but this may not be true on other datasets.

At evaluation time, we use beam search to explore a set of possible chains, which results in a set of chains $\mathbf{c}_1, \mathbf{c}_2, ... , \mathbf{c}_k$, with each chain containing different number of sentences.

\subsection{Answer prediction}
Since different beams may contain different plausible reasoning chains as shown in Figure~\ref{fig:overview}, we consider the sentences in the top $k$ beams predicted by our chain extractor as input to our answer prediction model. Different datasets may require different modifications of the basic BERT model as well as different types of reasoning, so we present the answer prediction module in the following section.

\section{Experimental Setup}\label{sec:exp}

\begin{table*}[t]
\small
\centering
\renewcommand{\tabcolsep}{1.3mm}
\begin{tabular}{c | c | c   c   c  c }
\toprule
Model  & Oracle & Avg Length  & Answer Found & Supp F1 & Answer F1   \\
\midrule
Oracle & Shortest  & 1.6 & 93.6  & 58.5  & - \\
Oracle & Q-Overlap  & 1.9 & 93.6 & 63.9 & - \\
Oracle & Supp Facts & 2.4 & 100.0 & 100.0 & 75.4 \\
\midrule
BERT-Para & Q-Overlap  & 2.0 & 76.3 & 64.5 & 66.0 \\
BERT-Para & Shortest  & 1.5 & 74.1 & 56.8 & 65.5 \\
BERT-Sent & Shortest & 1.7 & 72.5 & 53.1 & 60.2 \\
BiDAF-Para & Shortest  & 1.4 & 62.0 & 52.4 & 58.1 \\
\midrule
BERT-Para (top 5) & Q-Overlap  & 3.2 & 88.1 & 65.6 & \textbf{70.3} \\
\bottomrule

\end{tabular}
\caption{The characteristics of different chains generated by different models under different supervision on the HotpotQA dev set: for different models and chain oracles, we report the average chain length, fraction of chains containing the answer, F1 with respect to the annotated supporting facts, and F1 on the final QA task. Here we only pick the chain in the first beam.}
\vspace{-0.5cm}
\label{tab:chain_model_statistics}
\end{table*}

\subsection{Datasets}

\paragraph{WikiHop} \citet{welbl2018constructing} introduced this English dataset specially designed for text understanding across multiple documents. The dataset consists of around 40k questions, answers, and passages. Questions in this dataset are multiple-choice with around 10 choices on average.

\paragraph{HotpotQA} \citet{yang2018hotpotqa} proposed a new dataset with 113k English Wikipedia-based question-answer pairs. Similar to WikiHop, questions require finding and reasoning over multiple supporting documents to answer. Different from WikiHop, models should choose answers by selecting variable-length spans from these documents. Sentences relevant to finding the answer are annotated as ``supporting facts'' in the dataset.

\subsection{Implementation Details}
\paragraph{Oracle chain extraction}We use the off-the-shelf NER system from AllenNLP~\citep{Gardner2017AllenNLP}. We treat any entity that appears explicitly more than 5 times across sentences as a common entity,\footnote{These mentions are often extremely common entities like \emph{U.S.}, which are likely to introduce spurious edges rather than good ones.} and ignore it when we build the graph. Because these documents are only short snippets from Wikipedia, this criterion is loose enough to keep most useful mentions.

\paragraph{Chain extractor}We use the uncased BERT tokenizer to tokenize both question and paragraphs. We use the pretrained \texttt{bert-base-uncased} model and fine-tune it using Adam with a fixed learning rate of 5e-6. At test time, we produce our chains using beam search with beam size 5.  

\paragraph{Answer prediction} We concatenate the question and the combined chains from previous step in the top $k$ beams in the standard way as described in the original BERT paper~\cite{devlin2018bert} and encode it using the pre-trained BERT model. We denote its [CLS] token as $\textrm{[CLS]}_p$.

WikiHop is a multiple-choice dataset. Since we need to choose an answer from a candidate list, we encode each candidate with BERT. The [CLS] token for candidate $i$ is denoted as $\textrm{[CLS]}_{C_i}$. We then compute the score of a candidate $C_i$ being choose as the dot product between $\textrm{[CLS]}_p$ and $\textrm{[CLS]}_{C_i}$.

HotpotQA is a span-based question answering task, where finding the answer requires predicting the start and end of a span in the context. We compute distributions over these positions via two learned weight matrices $\mathbf{W}_\textrm{start}$ and $\mathbf{W}_\textrm{end}$. Each position in the concatenated sequence except the [CLS] token is multiplied by the corresponding weight matrix and softmaxed. Since we also need to predict the question type on HotpotQA (to handle yes/no questions vs. span extraction ones), we predict the type by taking the dot product of $\textrm{[CLS]}_p$ with a trainable matrix $\mathbf{W}_\textrm{type}$. We use \texttt{bert-large-uncased} instead of \texttt{bert-base-uncased} in the answer prediction module. We use the same optimizer and learning rate as chain extractor.

\section{Results}

In this section we aim to answer two main questions. First, which of our proposed chain extraction techniques is most effective, and how do they compare? Second, how does our approach compare to state-of-the-art models on these datasets? Finally, can we evaluate our extracted chains intrinsically: how important is ordering and how well do they align with human intuition about question answering?

\subsection{Comparison of Chain Extraction Methods}

In this section, we study the characteristics of our extracted chains with several experiments focused on HotpotQA. We choose this dataset since it provides human-annotated supporting facts so we can directly compare these against our model. Several statistics are shown in Table~\ref{tab:chain_model_statistics}. For different combinations of our model and which choice of chain oracle we use, we calculate several statistics, as described in the caption. We have the following observations:

\paragraph{Using more context helps chain extractors to find relevant sentences.} Comparing BERT-Para and BERT-Sent, we find that with all other parts fixed and only by encoding more context, we improve the answer prediction performance by around 5\%. This may indicate that BERT can capture cross sentence relations such as coreference and find more supporting evidence as a result. The comparison with BiDAF-Para vs. BERT-Sent also indicates this. Despite finding many fewer answer candidates (62\% instead of 72\%), BiDAF-Para only achieves around 2\% lower performance. One possible explanation to this is that without context, the BERT extraction model may pick up ``distractor'' sentences related to the question but do not actually lead to the answer, potentially confusing the answer prediction module.

\paragraph{The one-best chain often contains the answer.} This demonstrates the effectiveness of our chain extractor: the BERT-Para model with just 2 extracted sentences can locate the answer 76\% of the time. We further analyze the quality of these chains in the following sections. Note that this is nearly the same amount of evidence as in the human-labeled supporting facts (2.4 sentences on average); the difference can be explained by cases where the model can jump directly to the answer \citep{chen2019understanding}.

\paragraph{Q-Overlap helps recover more supporting evidence.} The main difference between our Shortest oracle and the Q-Overlap oracle is that Q-Overlap contains additional relevant sentences besides the one containing the answer. As a result, models trained with Q-Overlap should also yield a higher F1 score for finding the supporting facts, which is supported by the results (64 vs. 56).

\paragraph{Performance can be improved by taking a union across multiple chains} In the last row, we show a version of BERT-Para where the top 5 chains in the beam have been unioned together and truncated to 5 sentences. These top 5 chains contain permutations of roughly the same sentences, so this does not greatly increase the average length. However, this greatly increases answer recall and downstream F1. One reason is that this maintains uncertainty over the correct reasoning chain and can seamlessly handle question types involving comparison of multiple entities, which are difficult to address with a single reasoning chain of the sort presented in Figure~\ref{fig:overview}.

\subsection{Results compared to other systems}
We evaluate our best system from the prior section (BERT-Para with top-5 chains) on the blind test sets of our two datasets. Performance is shown in Table~\ref{tab:test_results}. On WikiHop, our model significantly outperforms past models, although these models do not use BERT. For HotpotQA, we use RoBERTa \citep{roberta} weights as the pretrained model instead of BERT, which gives a performance gain. Our model achieves strong performance compared to past models, including outperforming some models which use the human-labeled supporting facts~\footnote{This indicates that our heuristically-extracted chains can stand in effectively for this supervision, which suggests that our approach can generalize to settings where this annotation is not available.}

\begin{table}[t]
\small
\centering
\renewcommand{\tabcolsep}{1.3mm}
\begin{tabular}{ c | c   c  }
\toprule
   & dev & test \\ 
\midrule
GCN~\citep{de2018question} & 64.8 & 67.6 \\
BAG~\citep{cao2019bag} & 66.5 & 69.0 \\
CFC~\citep{zhong2019coarse} & 66.4 & 70.6 \\
JDReader~\citep{tu2019multi} & 68.1 & 70.9 \\
DynSAN~\citep{zhuang2019token} & 70.1 & 71.4 \\
BERT-Para (top 5) & \textbf{72.2} & \textbf{76.5} \\
\bottomrule
\end{tabular}
\hfill
\begin{tabular}{ c | c   c | c }
\toprule
   & EM & F1 & Supp? \\ 
\midrule
DecompRC~\citep{MinZZH19} & 55.20 & 69.63 & N \\
QFE~\citep{Nishida2019QFE} & 53.86 & 68.06 & Y\\
DFGN~\citep{Qiu2019DFGN} & 56.31	& 69.69	 & Y \\
HGN~\citep{fang2019hierarchical} & 66.07 & 79.36 & Y \\
SAE~\citep{tu2019select} & 66.92 & 79.62 & Y \\
Roberta-Para (top 5) & \textbf{61.20}	 & \textbf{74.11}  & \textbf{N} \\
\bottomrule
\end{tabular}
\caption{The blind test set performance achieved by our model on WikiHop and HotpotQA. On HotpotQA, all published works except DecompRC use the annotated supporting facts as extra supervision, which makes them not directly comparable to our model.}
\label{tab:test_results}
\end{table}

\begin{table*}[t]
\small
\centering
\renewcommand{\tabcolsep}{1.3mm}
\begin{tabular}{ c | c   c   |  c  c c | c c c }
\toprule
 Dataset & \multicolumn{2}{c|}{WikiHop} & \multicolumn{3}{c|}{HotpotQA} & \multicolumn{3}{c}{HotpotQA-Hard} \\ 
\midrule
 & Acc &  \%ans  & F1 & SP F1 & \%ans &  F1 & SP F1 &  \%ans  \\ \midrule
Chain Extraction & 72.4 & 72.7 & 69.7 & 63.7 & 90.3 & 56.0 & 59.2 & 78.7 \\
Unordered Extraction & 72.1 & 72.3 & 68.3 & 63.4 & 90.1 & 54.3 & 59.4 & 78.3 \\
\bottomrule
\end{tabular}
\caption{The downstream QA performance of the chains generated by different models on different datasets. The performance is evaluated by accuracy and F1 score respectively in WikiHop and HotpotQA dataset.}
\label{tab:performance_oracle_model}
\end{table*}

\begin{table*}[t]
\small
\centering
\renewcommand{\tabcolsep}{1.3mm}
\begin{tabular}{ c  c   c   c   }
\toprule
            & quite confident  & somewhat confident & not confident    \\
\midrule
shortest oracle  &  34 / 77.7 & 7 / 68.6  & 9 / 70.6  \\
extracted chain  & 37 / 81.1 & 7 / 64.2 & 6 / 50.0 \\
annotated supporting facts  & 33 / 78.8 & 12 / 60.0  & 5 / 88.0 \\
\bottomrule
\end{tabular}
\caption{The human evaluation on different evidence sets. For each row, 50 responses are bucketed based on the Turkers' confidence ratings, and numbers denote the answer F1 within that bucket.}
\vspace{-0.5cm}
\label{tab:human_eval}
\end{table*}



\subsection{Evaluation of chains}

\paragraph{Ordered extraction outperforms unordered extraction} One question we can ask is how important ordered chain extraction is versus just selecting ``chain-like'' sentences in an unordered fashion. We compare our BERT-Para model with a variant of our model where, instead of using a pointer network to predict a chain, we make an independent classification decision for each sentence to determine whether it is relevant to the question or not. We then pick top $k$ sentences with the highest relevance score and feed these to our BERT model. We name this model as \emph{unordered extraction}. Both are trained with the shortest-path oracle~\footnote{We do not use the question overlap oracle since the questions in WikiHop are synthetic like ``place\_of\_birth gregorio di cecco'', which is uninformative for the Q-overlap method.}. To make a fair comparison, we pick the same number of sentences ranked by prediction probability as the (top-5) chain extractor. 

QA performance on those datasets is shown in Table~\ref{tab:performance_oracle_model}. We also train and test our model on a hard subset of HotpotQA pointed out by~\newcite{chen2019understanding}. We see that \textbf{the sequential model is more powerful than the unordered model.} On all datasets, our chain extractor leads to higher QA performance than the unordered extractor, especially on HotpotQA-Hard, where multi-hop reasoning is more strongly required. This implies that even for a very powerful pre-trained model like BERT, an explicitly sequential interaction between sentences is still useful for recovering related evidences. A more powerful sequential decoder may further help with the those ''hard'' examples. On WikiHop, the improvement yield by our chain extractor is more marginal. One reason is that correlations have been noted between the question and answer options~\citep{chen2019understanding}, so that the quality of the extracted evidence contributes less to the models' downstream performance. 




\paragraph{Chain extraction is near the performance limit on HotpotQA} Given our two-stage procedure, one thing we can ask is: with a ``perfect'' chain extractor, how well would our question answering model do? We compare the performance of the answer prediction trained with our extracted chains against that trained with the human-annotated supporting facts. As we can see in Table~\ref{tab:chain_model_statistics}, BERT achieves a 75.4\% F1 on the annotated supporting facts, which is only 5\% higher than the result achieved by our BERT-Para (top 5) extractor. A better oracle or stronger chain extractor could help close this gap, but it is already fairly small considering the headroom on the task overall. It also shows there exist other challenges to address in the question answering piece, complementary to the proposed model, like decomposing the question into different pieces~\citep{MinZZH19} to further improve the multi-hop QA performance.

\begin{figure*}[t]
\centering
\includegraphics[width=150mm]{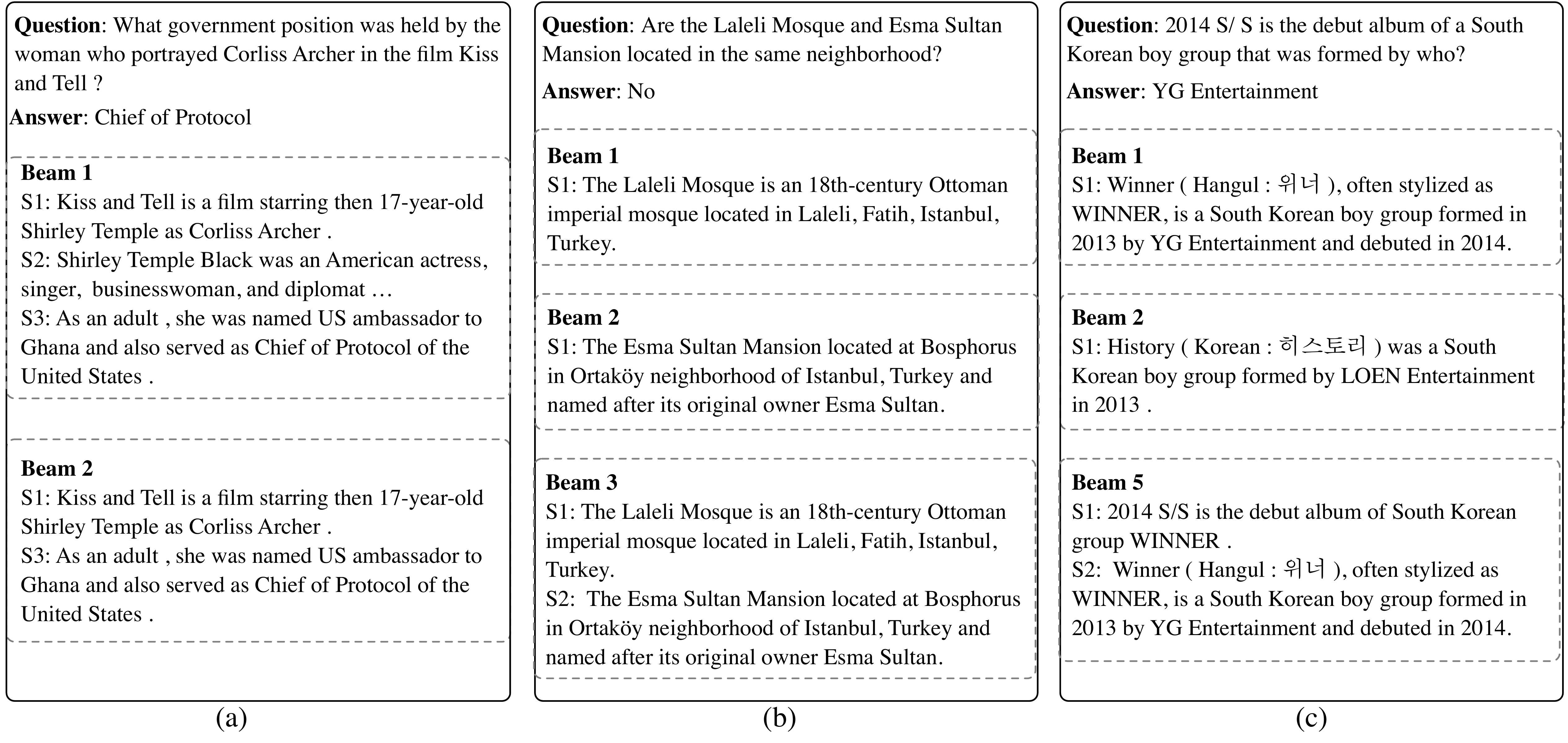}
\caption{Different chains picked up by our model on the dev set of HotpotQA. The first shows a standard success case, the second shows success on a less common question type, and the third shows a failure case.}
\vspace{-0.5cm}
    \label{fig:case_study}
\end{figure*}

\paragraph{Human evaluation}
We perform a human evaluation to compare the quality of our extracted chains with our oracle as well as the annotated supporting facts. We randomly pick 50 questions in HotpotQA and ask three Turkers to answer each question based on those different evidences and rate their confidence in their answer. We pick the Turkers' answer which has the highest word overlap with the actual answer (to control for Turkers who have simply misunderstood the question) and assess their confidence in it. The statistics are shown in Table~\ref{tab:human_eval}. Human performance on the three sets is quite similar -- they have similar confidence in their answers, and their answers achieve similar F1 score. Although sometimes the shortest oracle may directly jump to the answer and the extracted chains may contain distractors, humans still seem to be able to reason effectively and give confidence in their answers on these short chains. Since the difference between supporting facts and our oracle on overall question answering performance is marginal, this is further evidence that the human-annotated supporting facts may not be needed.

We also dig into the chains picked up by our chain extractor to better understand what is captured by our model. Those examples are shown in Figure~\ref{fig:case_study}. Seen from example (a), the model picks a perfect chain by first picking the sentence containing ``Kiss and Tell'' and ``Corliss Archer'', then finds the next sentence through ``Shirley Temple''. At the last step, to our surprise, it even finds a sentence via coreference. This demonstrates that although we do not explicitly model the entity links, the model still manages to learn to jump through entities in each hop.

Example (b) shows how our system can deal with comparison-style yes/no questions. There are two entities, namely, ``Laleli Mosque'' and ``Esma Sultan Mansion'' in the question, each of which must be pursued. The chain extractor proposes first a single-sentence chain about the first entity, then a single-sentence chain about the second entity. When unioned together, our answer predictor can leverage both of these together.

Example (c) shows that the extraction model picks a sentence containing the answer but without justification, it directly jumps to the answer by the lexical overlap of the two sentences and the shared entity ``South Korean''. The chain picked in the second beam is a distractor. There are also different distractors that contains in other hypotheses, of which we do not put in the example. The fifth hypothesis contains the correct chain. This example shows that if the same entity appears multiple time in the document, the chain extractor may be distracted and pick unrelated distractors.


\section{Related Work}

\paragraph{Text-based multi-hop reasoning}
Memory Network based models~\citep{weston2015towards,sukhbaatar2015end,kumar2016ask,dhingra2016gated,shen2017reasonet} try to solve multi-hop questions sequentially by using a memory cell which is designed to gather information iteratively from different parts of the passage. More recent work including Entity-GCN~\citep{de2018question}, MHQA-GRN~\citep{song2018exploring}, and BAG~\citep{cao2019bag}, form this problem as a search over entity graph, and adapt graph convolution network~\cite{kipf2017semi} to do reasoning. Such kind of models need to construct an entity graph both at training and test time, while we only need such entities during training.

\paragraph{Coarse-to-fine question answering}
\newcite{choi2017coarse} combine a coarse, fast model for selecting relevant sentences and a more expensive RNN for producing the answer from those sentences. \newcite{wang2019evidence} apply distant supervision to generate labels and uses them to train a neural sentence extractor. Another line of work proposes to use the answer prediction score as supervision to the sentence extractor \citep{wang2018r,indurthi2018cut,min2018efficient}. A recent line of works on open-domain multi-hop QA~\cite{feldman2019multi, das2019multi, asai2019learning, qi2019answering, godbole2019multi} also adopt the idea of forming query in a iterative way to select the most relevant documents regarding the question. Our model differs from those works for operating in a more fine-grind way: it actually shows how the answer is derived rather than just retrieves relevant documents. This represents a step towards building models that represent the reasoning process more explicitly \citep{trivedi2019repurposing, jiang2019explore}.

\section{Discussion and Conclusion}
In this work, we learn to extract reasoning chains to answer multi-hop reasoning questions. Experimental results show that the chains are as effective as human annotations, and achieve strong performance on two large datasets. However, as remarked in past work \citep{chen2019understanding,min2019compositional}, there are several aspects of HotpotQA and WikiHop which make them require multi-hop reasoning less strongly than they otherwise might. As more challenging QA datasets are built based on lessons learned from these, we feel that reasoning in a more explicit way and properties of chain-like representations will be critical. This work represents a first step towards this goal of improving QA systems in such settings.

\bibliography{emnlp-ijcnlp-2019}
\bibliographystyle{acl_natbib}

\end{document}